\documentclass[lettersize,journal]{IEEEtran}

\usepackage{amsmath,amsfonts}
\usepackage{algorithmic}
\usepackage{array}
\usepackage[caption=false,font=normalsize,labelfont=sf,textfont=sf]{subfig}
\usepackage{textcomp}
\usepackage{stfloats}
\usepackage{url}
\usepackage{verbatim}
\usepackage{graphicx}
\def\BibTeX{{\rm B\kern-.05em{\sc i\kern-.025em b}\kern-.08em
    T\kern-.1667em\lower.7ex\hbox{E}\kern-.125emX}}
\usepackage{balance}
\begin{document}
\title{GelSight FlexiRay: Breaking Planar Limits by Harnessing Large Deformations for Flexible, Full-Coverage Multimodal Sensing}
\author{Yanzhe Wang $^{1, *}$, Hao Wu $^{1, *}$, Haotian Guo $^{1, *}$, and Huixu Dong $^{1, \dag}$
\thanks{* Haotian Guo, Hao Wu, and Yanzhe Wang contribute equally to this work.}%
\thanks{1.\,Grasp Lab, Department of Mechanical Engineering, Zhejiang University, Hangzhou, China. {\small huixudong@zju.edu.cn}}%
}

\markboth{}%
{GelSight FlexiRay: Breaking Planar Limits by Harnessing Large Deformations for Flexible, Full-Coverage Multimodal Sensing}

\maketitle

\begin{abstract}
The integration of tactile sensing into compliant soft robotic grippers offers a compelling pathway toward advanced robotic grasping and safer human-robot interactions. Leveraging recent advances in computer vision, visual-tactile sensors realize high-resolution, large-area tactile perception purely with affordable cameras. However, conventional visual-tactile sensors rely heavily on rigid forms, sacrificing finger compliance and sensing regions to achieve localized tactile feedback. Enabling seamless, large-area tactile sensing in soft grippers remains challenging, as deformations inherent to soft structures can obstruct the optical path and restrict the camera’s field of view. To address these, we present Gelsight FlexiRay, a multimodal visual-tactile sensor designed for safe and compliant interactions with substantial structural deformation through seamless integration with Finray Effect grippers. First, we adopt a multi-mirror configuration, which is systematically modeled and optimized based on the physical force-deformation characteristics of FRE grippers. This design enables comprehensive surface perception using a single camera, even during large deformations. Second, we enhanced Gelsight FlexiRay with temperature-sensitive pigment markers, providing human-like multimodal perception, including contact force and location, proprioception, temperature, texture, and slippage. Experiments demonstrate Gelsight FlexiRay’s robust tactile performance across diverse deformation states, achieving a force measurement accuracy of 0.14 N and proprioceptive positioning accuracy of 0.19 mm. Compared with state of art compliant VTS, the FlexiRay demonstrates 5 times larger structural deformation under the same loads. Its expanded sensing area and ability to distinguish textures and temperature changes, monitor slippage, and execute grasping and classification tasks highlights its potential for versatile, large-area multimodal tactile sensing integration within soft robotic systems. This work establishes a foundation for flexible, high-resolution tactile sensing in compliant robotic applications.

\end{abstract}

\begin{IEEEkeywords}
multimodal sensing, vision-based tactile sensor, compliant gripper, optical optimization, neutral network.
\end{IEEEkeywords}

\section{Introduction}

\begin{figure}[ht]
    \centering
    \includegraphics[width=1\linewidth]{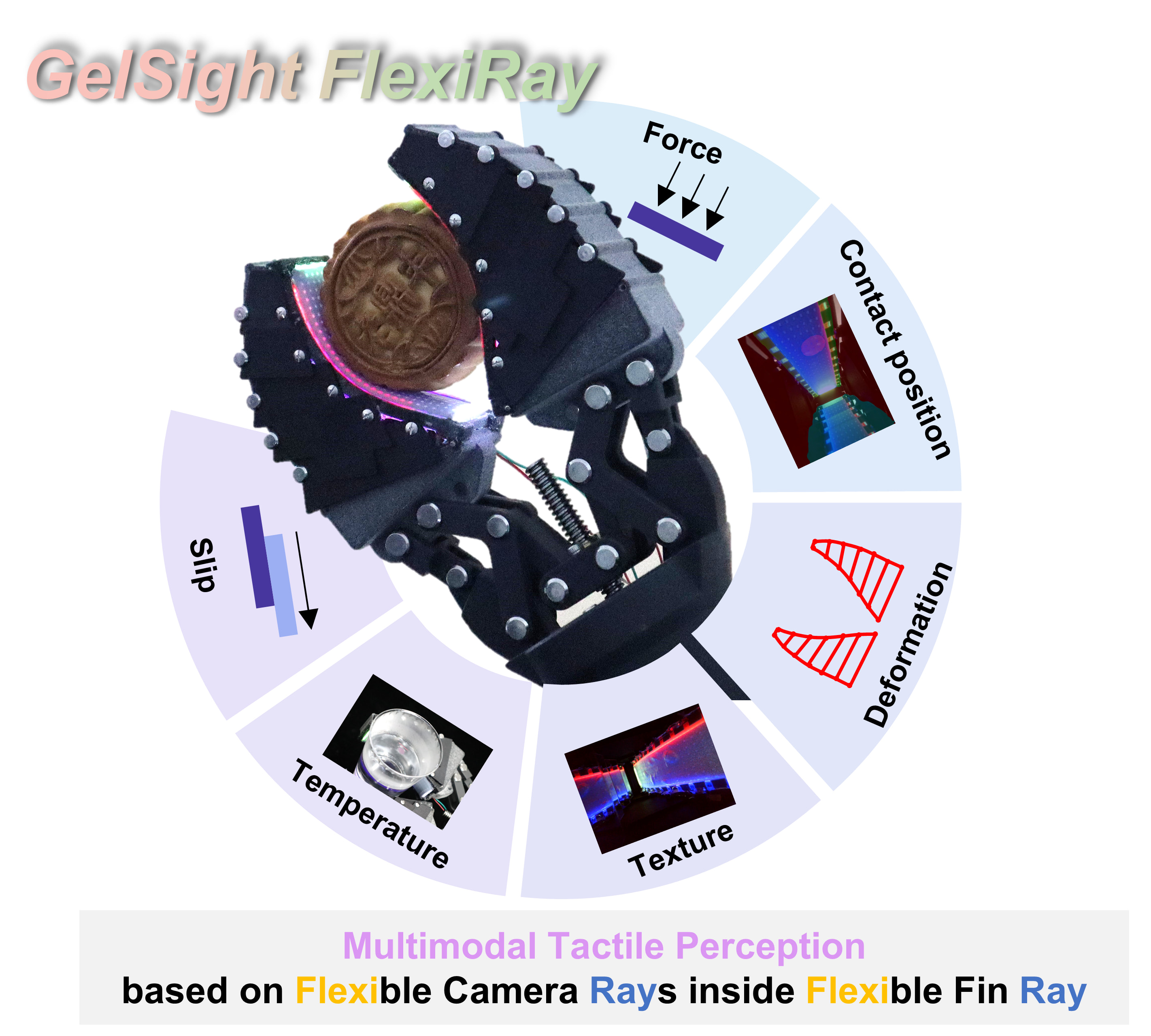}
    \caption{Gelsight FlexiRay: A novel flexible multimodal visual-tactile sensor, integrated with Fin Ray structures, delivers high-resolution tactile sensing on contact force, texture, slip, temperature, and proprioception.}
    \label{fig:titlepage}
\end{figure}

\IEEEPARstart{H}{uman} tactile perception is a highly complex and refined sensory ability, playing a crucial role in how humans perceive and understand the world\cite{handler2021mechanosensory}. This function can be categorized into three submodalities: the cutaneous system, the kinesthetic system, and the haptic system. The cutaneous system detects external physical contact and temperature stimuli through receptors located at various depths, such as mechanoreceptors and thermoreceptors. The kinesthetic system relies on receptors in muscles, tendons, and joints to convey information related to body posture. The haptic system integrates these two types of information to process the spatiotemporal characteristics of contact and interpret the feedback\cite{dahiya2009tactile}. Inspired by these sensory modalities, this study proposes the design of a novel soft robotic hand that replicates human tactile functions. The hand integrates "cutaneous," "kinesthetic," and "haptic" systems, featuring mechanoreceptors and thermoreceptors to sense external stimuli, provide feedback on hand posture, and interpret multidimensional contact information, including force, texture, slip, and temperature.

Compared to rigid grippers\cite{dong2018geometric}, soft grippers offer enhanced compliance and adaptability\cite{dong2021gsg}, enabling them to naturally conform to the shapes of various objects. This improves the robot's robustness and safety in complex, uncertain environments \cite{shintake2018soft, waveguide_integrated}. In this study, we utilize a bio-inspired soft gripper based on the Fin Ray structure. The Finray effect (FRE) allows for adaptive contact through passive, compliant deformation. It can be actuated using a simple opening and closing mechanism, making it compatible with existing industrial equipment without the need for a complex control system. This low-cost, efficient soft structure provides an ideal foundation for the integration of tactile sensors. However, integrating tactile sensors with the soft gripper presents significant challenges. First, existing tactile sensors and soft structures are often developed independently, leading to compatibility issues during integration\cite{zou2017novel, wang2018skin, waveguide_integrated}. The rigid structure of conventional sensors can compromise the compliance of the soft hand. Second, while flexible electronics offer promising solutions for tactile sensing, achieving high-performance perception often requires more complex electronics and higher costs, making large-area, high-resolution coverage difficult to achieve\cite{wang2015recent, wang2018skin}. Third, enabling multimodal tactile sensing, akin to human touch, is a critical challenge, particularly in efficiently capturing and interpreting multidimensional information from multiple stimuli under flexible conditions.

Vision-based Tactile Sensors (VTS) have emerged as a promising solution to address these challenges, offering advantages such as low cost, high resolution, and multimodal sensing capabilities\cite{yuan2017gelsight, fang2024force}. VTS capture pixel-level tactile feedback through cameras, enabling perception that closely approximates human tactile sensing\cite{kakani2021vision, yuan2018active}. However, integrating VTS with soft grippers still faces significant technical challenges. First, VTS typically rely on rigid structures to maintain the stability of the optical system, which conflicts with the flexible nature of soft grippers. Second, traditional VTS are often limited in their sensing range, typically confined to the fingertip area, and lack designs that provide full coverage of the soft gripper's contact surface. Furthermore, the perceptual field of existing VTS is generally restricted to a two-dimensional(2D) plane. When the soft gripper undergoes bending deformation, the contact surface dynamically changes in three-dimensional space(3D), potentially leading to camera occlusion and significantly limiting the ability to perceive non-planar contact surfaces. These issues constrain the deep integration and practical deployment of VTS in soft grippers.

Motivated by the unresolved challenges, this study introduces FlexiRay, a novel FRE gripper that integrates VTS with a compliant, soft-material architecture, enabling flexible, full-coverage multimodal and proprioceptive sensing. To address the gap in integrating tactile sensing with soft grippers, we adopt a collaborative design strategy that unifies perception and the underlying structure. The overall architecture is based on the modal hierarchy of the human hand. The flexible back beam of the FRE gripper acts as muscles and tendons, while the rigid side beams serve as the bones of the hand, with articulated joints at both ends providing deformable motion. The front beam is constructed with a flexible polydimethylsiloxane(PDMS) substrate and soft silicone to form the cutaneous. The contrast in material stiffness enables interaction-induced deformations, acting as mechanoreceptors sensitive to external contact. Embedded within the cutaneous are temperature-sensitive pigments that function as thermoreceptors. The camera, serving as the central nervous system, encodes the sensor responses and proprioceptive joint information, which are then processed by a learning model to decouple and interpret the tactile sensory data.

In addressing the compatibility challenges between VTS and the flexible structure, the key issue arises from the significant structural deformations during interaction. These deformations alter the optical path, obstructing the camera’s field of view(FOV) of the contact areas, which restricts the VTS to rigid structures and a limited 2D sensing plane. In contrast, rather than limiting the flexibility of the structure\cite{liu2022gelsight}, we leverage these deformations as an advantage. Through systematic optimization, we introduce a multi-mirror configuration that exploits large deformations. Each mirror independently follows the passive movement of the tendons to adjust the direction of tactile signal transmission, allowing discrete capture of areas not covered by the camera’s FOV. The combination of multiple mirrors’ FOV enables continuous coverage of a large sensing area. Finally, based on the optimized soft gripper, we refine the partitioning of the cutaneous and kinesthetic systems using a semantic segmentation model. Using a data-driven learning model, we decouple the proprioceptive sensing of contact force, position, and joint location. Then, multimodal sensing models are developed, which integrate with specific tasks and explores its potential in complex real-world applications.

We emphasize the contributions of this work.
\begin{enumerate}
    \item A novel integrated design for VTS and FRE soft grippers is proposed, realizing a compact, humanoid-inspired tactile sensing modality with a hierarchical structure that balances compliance and sensing performance.

    \item A new flexible VTS substrate architecture is developed, combining an integrated manufacturing process with the FRE base structure, a PDMS-based contact base, and a flexible silicone touch material. A temperature-sensitive material are also incorporated to enhance the multimodal sensing capabilities.

    \item A layout optimization method for the inside optical sensing system, based on Covariance Matrix Adaptation Evolution Strategy (CMA-ES), is proposed. This method leverages physically induced adaptive deformation to optimize the configuration of a single camera and multiple mirrors, maximizing perception coverage under deformation. It transforms optical interference from a limiting factor into a functional design element, enabling large-scale deformation and extensive VTS coverage.

    \item In the flexible 3D spatial perception domain, FlexiRay achieves a force sensing accuracy of 0.14 N and a proprioceptive positioning accuracy of 0.19 mm, while simultaneously enabling multimodal tactile capabilities such as texture recognition, temperature perception, and slip detection. This offers a promising solution for advanced soft robotic applications.

\end{enumerate}

\section{Related Work}

\subsection{Sensorized Soft Gripper}

Sensorized soft gripper are playing an increasingly important role in grasping, manipulation, and physical human-robot interaction tasks, as they provide robots with both mechanical compliance and simultaneous tactile feedback \cite{waveguide_integrated, acoustic}. Significant advances have enriched the sensing capabilities of soft robotic hands, encompassing tactile, proprioceptive, and multimodal perception leveraging principles such as resistive \cite{Science_resistive, zhou2022sensory}, magnetic \cite{magnetic}, pressure-sensitive \cite{soft_pressure, soft_pressure_icra}, capacitive \cite{zhou2022sensory}, waveguide-based \cite{waveguide_integrated, waveguide_pri}, acoustic \cite{acoustic}, and thermal sensing \cite{heat} et al. Taxel-based measurement methods, in particular, are known for their high accuracy and sensitivity. For instance, inspired by the mechanoreceptive properties of Merkel cells and Ruffini endings in human skin, Liu et al. developed a 3D-architected electronic skin capable of precisely decoupling normal force, shear force, and strain, demonstrating excellent multimodal sensing fidelity \cite{Science_resistive}.
However, these taxel-based approaches are often limited by resolution, higher costs, and structural fragility.

With the rise of data-driven, learning-based approaches, "computational sensors" are emerging as a compelling alternative, where raw signals are processed to infer tactile properties beyond direct sensor outputs, liberating design from hardware-specific constraints \cite{acoustic, spiegel}. A variety of novel sensing principles, including sound, light, and heat, have been applied in soft robotic systems based on "computational sensors". A key design goal of these methods is to achieve broad coverage with a minimal number of sensors. For example, Meerbeek et al. utilized an internally illuminated elastomer foam, capturing intensity changes from embedded optical fibers to achieve exceptional accurate proprioceptive sensing of bending and twisting, with a mean absolute error of only 0.06° \cite{waveguide_pri}. Similarly, Wall et al. introduced acoustic sensing in the soft pneumatic fingers, with a speaker and microphone embedded within the chamber, enabling multi-modal sensing of contact location, force, and temperature \cite{acoustic}. Li et al. exploited thermal conductivity changes in porous materials under varying contact conditions, enabling simultaneous sensing of material thermal properties, pressure, and ambient temperature \cite{heat}. Harnessing massive data, these systems achieve multimodal, large-area perception with minimized sensors. Yet, they face limitations in spatial resolution, as exemplified by Wall et al.'s hand, which detects only six discrete contact locations and three force levels \cite{acoustic}. To fully realize the potential of soft robotics in complex, adaptive tasks, these designs must navigate trade-offs between resolution, coverage, cost-effectiveness, and multimodal sensing. A comprehensive, human-like tactile modality solution remains challenging and elusive.

\subsection{Vision-based Tactile Sensing}

The visual-tactile sensor, introduced by Johnson et al. \cite{CVPR_gelsight_grandpa}, represents a groundbreaking method for acquiring rich tactile information from gel-like contact surfaces using cameras. Unlike other sensing principles, VTS uniquely provides large-area, high-resolution, pixel-level sensory information using cost-effective cameras, making it appealing for diverse robotic applications \cite{fang2024force, abad2020visuotactile}. Typically integrated into robotic fingertips or palms, VTS enables localized sensing of force, contact location, and texture. Notable examples include GelSight \cite{yuan2017gelsight} and Digit \cite{lambeta2020digit}. However, due to optical path constraints, traditional VTS designs tend to be bulky and have limited sensing coverage. Efforts have been made to optimize optical paths via camera layout adjustments or the incorporation of plannar mirrors. For instance, Donlon et al. developed GelSlim, which utilizes a planar mirror to create a compact optical path, greatly reducing the sensor's overall size. Other approaches, such as GelSight 360 \cite{tippur2023gelsight360} and DenseTact \cite{do2022densetact}, position the camera beneath a curved elastic surface, making full utilization of the wide field of vision of cameras, to achieve omni-directional force and contact sensing. Besides, VTS is inherently multimodal, allowing the realization of pixel-level force sensing, proximity detection, 3D surface reconstruction, and even super-resolution temperature sensing, which significantly expands it potential application \cite{fang2024force, xu2024vision}. For example, Abad et al. adopt a conventional GelSight design with thermochromic materials to enable rapid temperature detection, based on which they achieve human-inspired withdrawal reflex \cite{abad2021haptitemp}. A step further, Li et al. developed the M$^{3}$Tac sensor, which integrates visible, near-infrared, and mid-infrared imaging to achieve high-resolution sensing of deformation, texture, force, stickiness, temperature, and proximity, way surpassing human skin in sensory performance \cite{10682561}. However, in summary, most of these sensors are designed as standalone units, which complicates their seamless integration into robotic grippers. This underscores the need for holistic designs that combine VTS capabilities with the compliance and adaptability of robotic systems.

Improving the integration of VTS with robotic hands is crucial for enhancing perception coverage and durability. For instance, Zhao et al. designed GelSight Svelte, which incorporates a single curved mirror inside the hand to expand the sensing area, achieving a compact, single-camera VTS comparable in size to a human finger \cite{zhao2023gelsight}. Similarly, Sun et al. systematically optimized the design of omni-directional VTS by combining an improved hollow-structured skeleton with an elastomeric surface coating, increasing localized compliance \cite{sun2022soft}. However, these designs still rely on rigid underlying structures, where the structural rigidity prevents full utilization of the expanded sensing area during real-world interactions. To address this limitation, She et al. develop GelFlex, a cable-driven soft gripper incorporating GelSight-like architectures. Employing a learning-based approach, it achieves both proprioception and shape classification, even under large deformations \cite{she2020exoskeleton}. However, significant occlusion occurs during large deformations, necessitating the use of multiple cameras for segmented coverage, which increases both cost and maintenance complexity. Liu et al. makes an attempt to integrate GelSight architecture into biologically inspired compliant FRE grippers to enable high-resolution tactile sensing during adaptive grasping \cite{liu2022gelsight, liu2023gelsight}. Their approach involves replacing rigid acrylic with thin, flexible mylar at the front contact surface to enhance deformation. However, the back beams and crossbeams of the FRE grippers remain rigid to preserve the optical path, failing to address the inherent limitations in the compliance of current VTS designs. Concurrently, this rigidity compromises the natural adaptability of the FRE mechanism, diminishing its key advantage. As a result, current VTS implementations continue to exhibit suboptimal performance in interactions with non-flat surfaces, falling short in achieving high-resolution, large-coverage, and multimodal sensing during compliant interactions. Addressing these challenges necessitates innovative solutions that enable conformal grasping while ensuring robust sensing efficiency, even under significant deformation and complex interaction scenarios.

\section{Design and Fabrication}

This section details the design and fabrication process of the proposed GelSight FlexiRay. It begins with an introduction to the mechanical design and the internal component layout. Next,a parametric modeling approach is presented alongside a CMA-ES-based optimization method. This method enhances the sensing coverage of the optical path under different deformations in the spatial geometry of the camera and multiple mirrors. Finally, the manufacturing and production process of the FlexiRay is elaborately described.

\subsection{Mechanical Design}

\begin{figure*}[ht]
    \centering
    \includegraphics[width=1\linewidth]{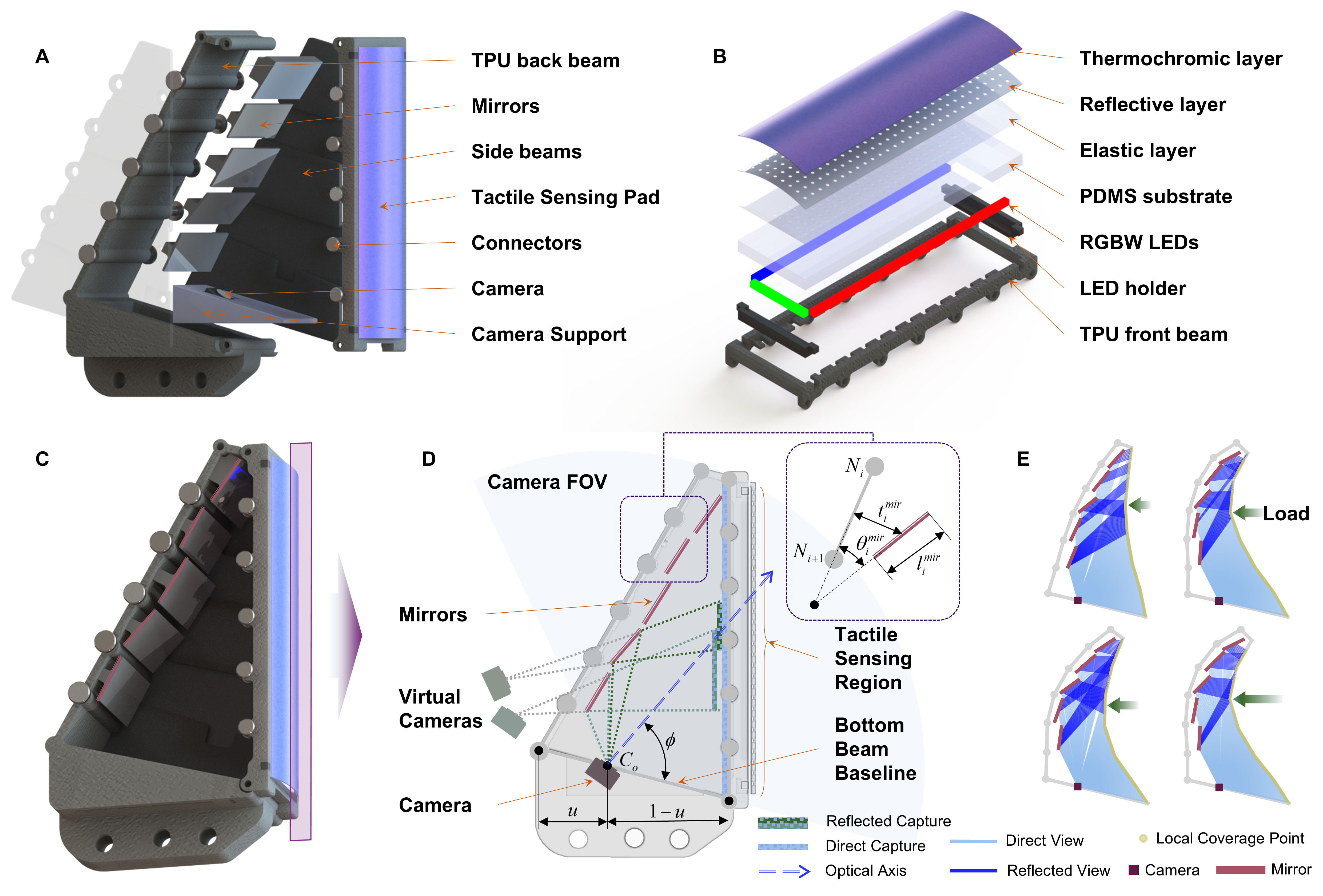}
    \caption{Illustration of the integrated design and layout optimization of Gelsight FlexiRay. (A) Exploded view of Gelsight FlexiRay. (B) Exploded view of the tactile sensing pad. (C) 3D perspective of the multi-mirror layout inside Gelsight FlexiRay. (D) Schematic representation of optical system optimization principles in a 2D cross-sectional view. (E) Camera FOV coverage under different deformation conditions due to varying loads.}
    \label{fig:exploded}
\end{figure*}

The proposed Gelsight FlexiRay design is illustrated in Fig.\ref{fig:exploded}(A) and (B). It consists of five main components: an optical camera for image capture, a complaint finger framework, reflective mirrors for visual enhancement, LED light strips and light-blocking beams for illumination, as well as a layer of silicone gel for tactile surface interaction.

\begin{itemize}
    \item Camera. We chose the camera with high-resolution image capture capability (12 million pixels), compact structure (8 x 8.5 x 5 mm), and wide angle (135°). Hence, the camera can be positioned within the gripper to obtain clear internal imagery.
    \item Finger. The Fin Ray finger manufactured using TPU materials is capable of conforming to various object shapes without active actuation.
    \item Mirrors. The mirrors are strategically mounted on the finger's backbone at specific angles and positions to augment the visual field, ensuring comprehensive visibility even during substantial deformations.
    \item Illumination. Flexible RGBW (red, green, blue, white) LED lights are embedded in the silicone gel in a series configuration during fabrication. Interlaced side beams are designed to prevent environmental light interference while preserving adaptability.
    \item Tactile Sensing Pad. The tactile sensing pad primarily consists of four layers. We combine Silicone Inc. 00-30 silicone with thermochromic pigment to form the temperature-sensing layer. For the reflective layer, silicone is mixed with aluminum flake and aluminum powder to enhance reflectivity. To enable the acquisition of detailed textural information, a low-hardness (5 A) transparent silicone is used as the elastic layer. Additionally, we employ PDMS as an elastic support to preserve the compliance and deformation capacity of the Fin Ray.
\end{itemize}

% 插图说明用软材料（如PDMS）替换传统亚克力，对于接触的影响，并加入公式定性分析。

\subsection{CMA-ES based Optics Layout optimization}

The equilibrium between finger compliance and the robustness of the tactile sensing region presents a significant challenge in the advancement of soft VTS. To mitigate the limitations of FOV and occlusion blind spots associated with large-deformation finger structures, a multi-mirror FOV optimization method is proposed, leveraging CMA-ES\cite{hansen2003reducing, nomura2024cma}. Through the optimization of both structural and layout parameters of the camera and mirrors, a visual reflection system composed of multiple planar mirrors is established on the back side of the finger. This configuration facilitates a single camera in achieving comprehensive coverage of the entire deformation range and contact region of the Fin Ray, thereby enabling enhanced visual-tactile perception.

Initially, the displacement of nodes under the applied force \( F \) is collected for the Fin Ray, which has identical structural parameters. The displacement is represented by the coordinates of the lateral 2D cross-section, with the back joint nodes denoted as \( \{ N_i \}_{i=1}^{n} \) and the tactile sensing area joint nodes labeled as \( \{ P_i \}_{i=1}^{n} \). Consequently, a mapping from force to deformation is established, represented as \( f: F \rightarrow \{ N_i, P_i \}_{i=1}^{n} \).

To preserve the compliance of Fin Ray without introducing structural interference, the rigid planar mirror is designed in a T-shape. This configuration minimizes the bonded area with the back flexible beam while ensuring that the reflective surface remains sufficiently large, as illustrated in Fig.\ref{fig:exploded}(C-D). The subsequent part provides a detailed account of the construction and solution of the optical layout optimization problem.

\textbf{Decision variables:}
\begin{equation} 
\begin{split}
x = & \left\{ (\theta_i^{mir}, t_i^{mir}, l_i^{mir}) \right\}_{i=1}^{n-1} \cup \{ u, \phi \} \\
    & \quad \text{with} \quad \theta_i^{mir}, t_i^{mir}, l_i^{mir}, u, \phi \in \mathbb{R} 
\end{split}
\end{equation}
where \( \{ \theta_i^{mir}, t_i^{mir}, l_i^{mir} \} \) represents the parameters associated with the mirrors. Specifically, \( \theta_i^{mir} \) denotes the angle between the mirror and the line segment formed by two adjacent nodes on the back side, \( t_i^{mir} \) indicates the midpoint offset distance, and \( l_i^{mir} \) signifies the length of the mirror. The camera is positioned along the baseline defined by the back side and the base nodes of the tactile sensing pad. The distance from the camera to the back base node is expressed as a coefficient \( u \), relative to the length of the baseline. Furthermore, the angle between the optical axis of the camera and the baseline is represented by \( \phi \).

\textbf{Objective Function:}

Given a camera with a fixed FOV, the discrete rays produced by the camera can be represented as \( \mathcal{R} = \{r_j\}_{j=1}^{m} \). The primary objective of the optimization process is to maximize the coverage of the tactile sensing region by the collection of FOV rays across various deformations. This goal can be achieved through two approaches: direct imaging and single reflections from mirrors.

By sampling \( K \) distinct loads from the force-displacement mapping \( f \), a set of joint nodes \( \mathcal{D} \) is obtained. Each deformation structure \( d \) corresponds to a discrete set of target points, denoted as \( \mathcal{P}^d \), within the tactile sensing region. The radius of the illumination range for a single FOV ray is defined as \( R \propto l^c \), which depends on the propagation distance \( l^c \). Consequently, the objective function can be expressed as follows:
\begin{equation}
\text{Maximize } f(x) = \frac{1}{K} \sum_{d \in \mathcal{D}} \sum_{r \in \mathcal{R}} \sum_{p \in \mathcal{P}^d} I(x, r, p, R)
\end{equation}
where \( I(x, r, p, R) \) is an indicator function that returns 1 if the target point resides within the coverage range of the ray beam; otherwise, it returns 0.

\textbf{Constraints:}

The following constraints must be adhered to:

\begin{itemize}
    \item \textbf{Geometric Constraints:} The lengths, offsets, and rotation angles of each mirror, as well as the position of the camera, must remain within predefined ranges:  \( l_i^{mir} \in [l_{\text{min}}, l_{\text{max}}], \, t_i^{mir} \in [t_{\text{min}}, t_{\text{max}}], \, \theta_i^{mir} \in [\theta_{\text{min}}, \theta_{\text{max}}], \, u \in (0, 1). \)

    \item \textbf{Occlusion Constraints:} The potential occlusion between mirrors during the deformation process must be taken into account. If an occlusion exists along the line of sight, the indicator function \( I(x, r, p, R) \) returns 0.

back    \item \textbf{Safety Constraints:} Under all deformation conditions, mirrors must not interfere with the front and back beams. If this condition is not met, the indicator function \( I(x, r, p, R) \) returns 0 as a penalty.
\end{itemize}

\textbf{Solution Process:}

The optimization process utilizes CMA-ES algorithm\cite{nomura2024cmaes} to search for optimal layout solutions by sampling candidate solutions from a multivariate Gaussian distribution. The multivariate Gaussian distribution is defined as \( \mathcal{N}(\mu, \sigma^2 C) \), where \( \mu \) is the mean vector, \( C \) is the covariance matrix, and \( \sigma \) represents the step size.

In the \( y \)-th generation of the optimization process, for a population size of \( \beta \), the candidate solutions \( \{ x_i \}_{i=1}^{\beta} \) are obtained by sampling from the distribution \( \mathcal{N}(\mu^{(y)}, (\sigma^{(y)})^2 C^{(y)}) \):
\begin{equation}
x_i = \mu^{(y)} + \sigma^{(y)} \sqrt{C^{(y)}} s_i
\end{equation}
where \( s_i \sim \mathcal{N}(0, I) \), with \(I\) denoting the identity matrix. The \( k \) solutions with the highest objective function values are selected as candidates: \(\{ f(x_{1:k}) \mid f(x_1) \geq f(x_2) \geq \dots \geq f(x_k) \geq \dots \geq f(x_{\beta}) \}\)

The evolution paths are updated using the following equations:
\begin{equation}
p_\sigma^{(y+1)} = (1 - c_\sigma) p_\sigma^{(y)} + \sqrt{c_\sigma (2 - c_\sigma) \lambda_w}  \sqrt{ C^{(y)} }^{-1} \delta
\end{equation}
\begin{equation}
p_c^{(y+1)} = (1 - c_c) p_c^{(y)} + \sqrt{c_c (2 - c_c) \lambda_w} H_\sigma^{(y+1)} \delta
\end{equation}
where \( c_\sigma \) and \( c_c \) are cumulation factors, respectively. \( H_\sigma^{(y+1)} \) is the Heaviside function, and
\[
\delta = \sum_{i=1}^{\eta} w_i \sqrt{\Sigma^{(y)}} s_i,\quad\lambda_w = \frac{1}{\sum_{i=1}^{\eta} w_i}
\]

Therefore, the parameters of the multivariate Gaussian distribution for each generation are updated as follows:
\begin{equation}
\mu^{(y+1)} = \mu^{(y)} + c_l \sigma^{(y)} \sqrt{C^{(y)}} s_i
\end{equation}
\begin{equation}
\sigma^{(y+1)} = \sigma^{(y)} \exp\left( 1, \textstyle {c_\sigma \left( \frac{\| p_\sigma^{(y+1)} \|}{\mathbb{E}[\mathcal{N}(0,I)]} - 1 \right) }/{d_\sigma}\right)
\end{equation}
\begin{multline}
C^{(y+1)} = \left(1 + c_1 c_\mu (2 - c_c)(1 - H_\sigma^{(y+1)})\right) C^{(y)}\\
+ c_1 \Delta_1 + c_\mu \Delta_\mu
\end{multline}
where \(\Delta_1\) and \(\Delta_\mu\) correspondingly denote the rank one and \(-\mu\) updates. The parameters \(c_l\), \(c_1\), and \(c_\mu\) represent the learning rates for the mean, rank one, and \(-\mu\) updates, respectively. The damping factor \(d_\sigma\) is used for the adaptive accumulation of step sizes.

Starting with initial parameters that satisfy the constraints, the parameters of the multivariate Gaussian distribution are updated iteratively. This process continues until either the expected stopping condition for the objective function is met or the maximum number of optimization generations is reached. The optimized camera and mirror layout parameters are then derived from the distribution.

\subsection{Fabrication and Manufacturing}

\begin{figure*}[ht]
    \centering
    \includegraphics[width=1\linewidth]{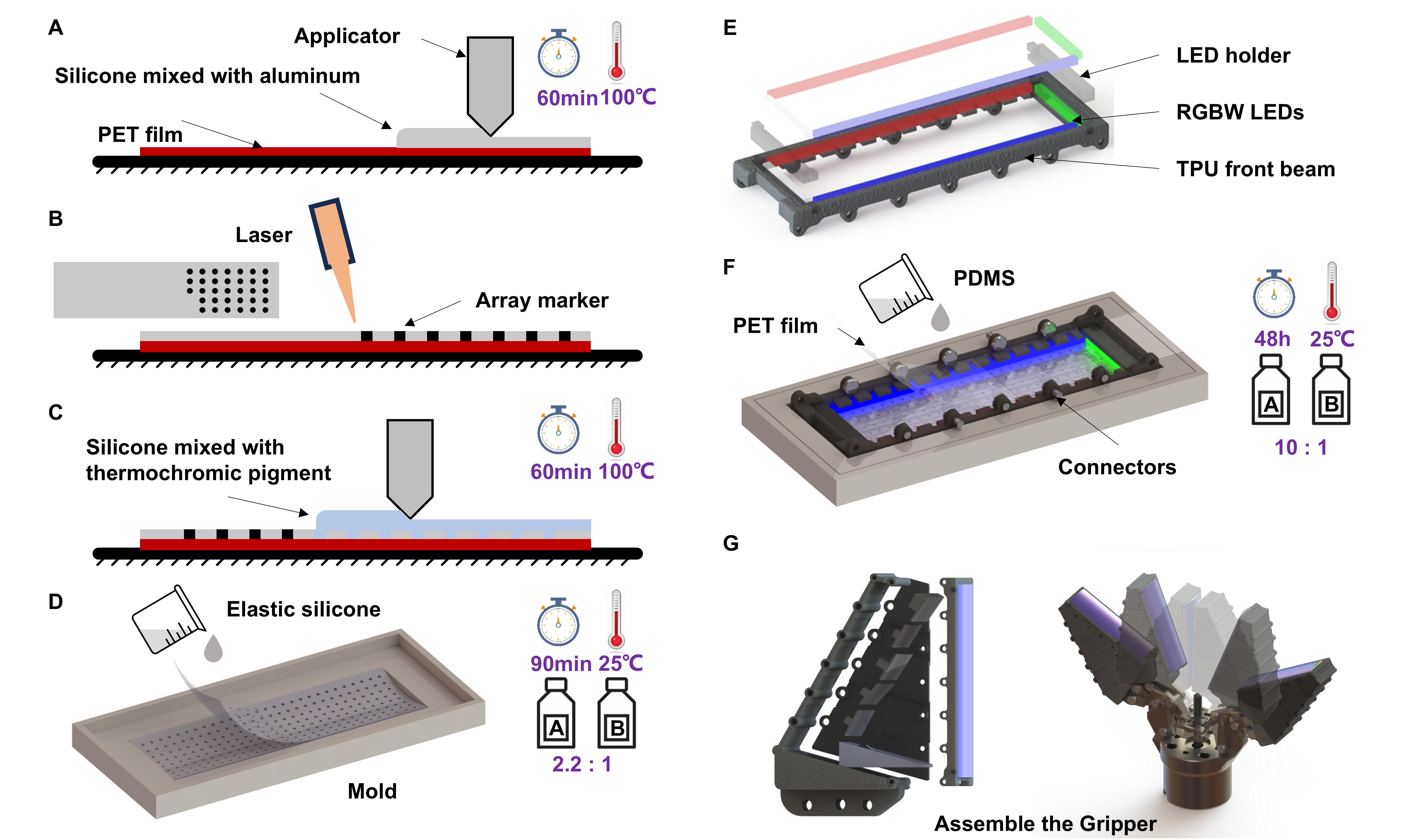}
    \caption{Fabrication process of Gelsight FlexiRay. (A) Casting of the reflective layer. (B) Engraving markers with a laser. (C) Casting of the temperature sensing layer. (D) Casting of the soft elastic layer. (E) String the LED lights in series and mount them onto the TPU beam. (F) Casting of the PDMS support. (G) Assemble the remaining 3D-printed parts to form the gripper.}
    \label{fig:fabrication}
\end{figure*}

Our primary objective is to endow Fin Ray with the capability to perceive tactile information through a vision-based tactile method, while preserving their compliance and passive deformation abilities. To achieve this, we have replaced the tradition acrylic supporter with PDMS material and opted for flexible LED lights for illumination. The fabrication process is as follows:

Firstly, Silicone Inc. 00-30 silicone is mixed with aluminum powder and flake in a mass ratio of 400:20:3 as the reflective layer. This mixture is spread with an adjustable squeegee applicator to create a silicone film with a thickness of 0.15 mm, which is then cured at 100 °C for 1 hour. Subsequently, a laser cutter is utilized to remove an array of round markers. The silicone is again mixed with two thermochromic pigments at a mass ratio of 10:1:1 as the temperature sensing layer. The powders possess thermochromic properties, enabling them to detect temperature variations and undergo color changes in response to temperatures exceeding or falling short of the threshold values. Hence, the two thermosensitive powders facilitate a temperature-responsive transition of the thermometric layer: it exhibits a deep red hue at temperatures above 38°C, shifts to a deep blue below 18 °C and appears as a light purple under ambient conditions. With the same application method, we can obtain the 0.15 mm thick temperature sensing layer and bond the two layers together. Afterwards, a prepared composite layer is placed into the mold, and a low transparent silicone is mixed at a ratio of 2.2:1 to achieve a soft elastic layer with 5 shore A hardness, similar to the hardness of human skin. The silicone is poured into the mold and allowed to cure at room temperature for 90 minutes. Subsequently, the RGBW leds are connected in series and affixed to the TPU front beam. PDMS is mixed in a ratio of 10:1 to serve as a transparent support that allows for large deformations. The TPU front beam and LED lights are placed into the mold. Further, PDMS material is poured to encapsulate them within, resulting in a tactile sensing pad capable of perceiving texture, temperature, and other information.

After fabricating the tactile sensing silicone pad, we can integrate the mirrors, camera, side beams, and finger framework with the pad to construct a Fin Ray finger with the multimodal sensing capability. To validate its perception and grasping capabilities, we propose a gripper comprising two identical Fin Ray fingers. A single stepper motor actuates the gripper through linkage mechanisms, enabling a substantial range of motion for opening and closing actions.

\subsection{Perception Models}

\begin{figure*}[ht]
    \centering
    \includegraphics[width=1\linewidth]{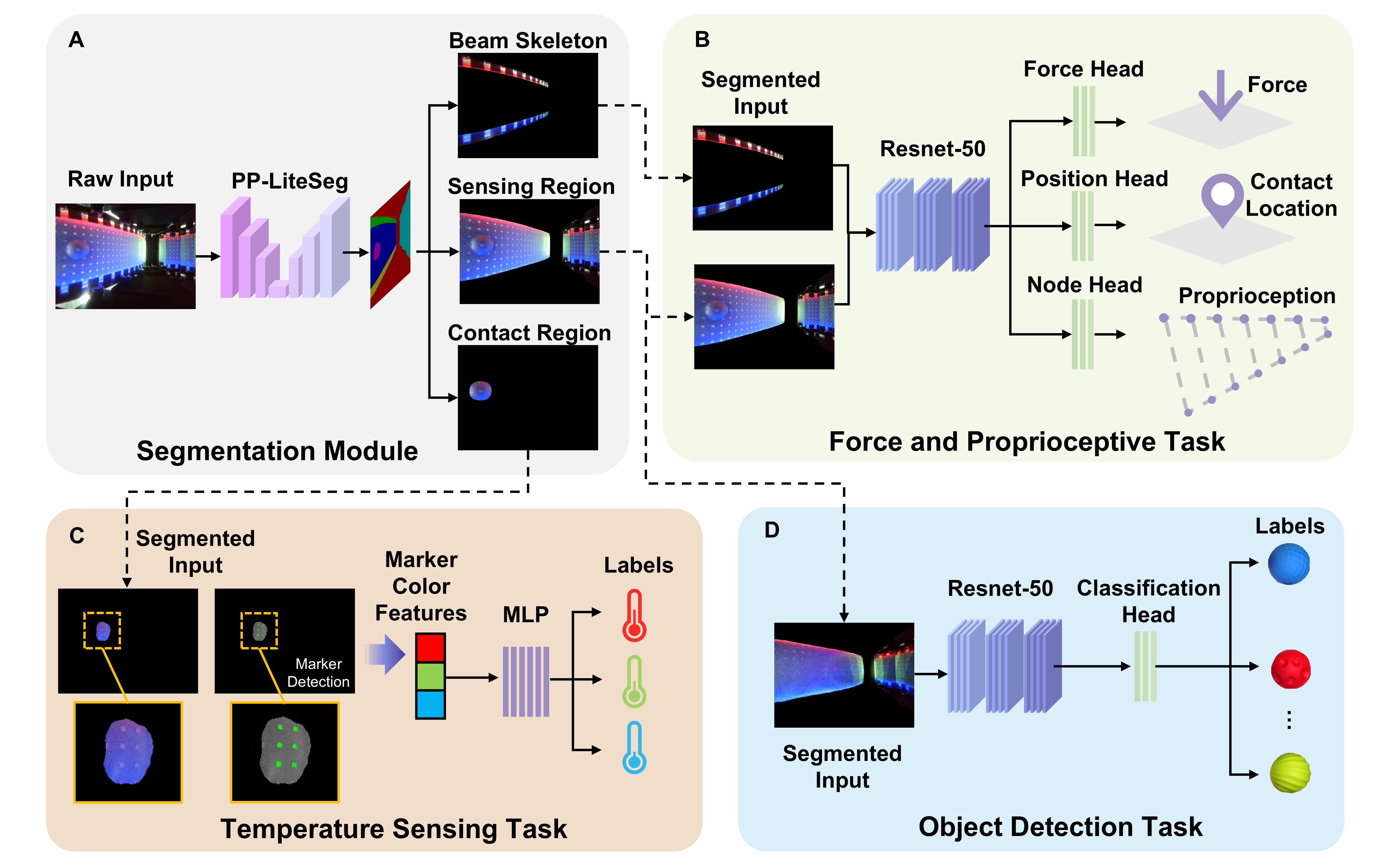}
    \caption{Perception model architecture. (A) Real-time semantic segmentation model for segmenting the front beam skeleton, perception region, and contact region. (B) Regression model for sensing interaction normal force, position, and proprioceptive structure deformation within the perception region. (C) Marker color mapping model for temperature sensing in the contact region. (D) Tactile-based object classification model for texture perception and recognition.}
    \label{fig:model}
\end{figure*}

Traditional VTS typically rely on rigid structures, where the tactile sensing area remains fixed within the camera’s FOV. Image changes in these systems are solely induced by contact. By acquiring an initial image under non-contact conditions and performing image differencing, it becomes possible to capture contact regions and other features. However, the flexible nature of the Fin Ray structure introduces both intrinsic and contact-induced deformations during interaction. This shifts the problem from a 2D to a 3D visual challenge. As a result, traditional VTS methods are no longer applicable, and interpreting tactile information becomes significantly more complex. To address this issue, a learning-based approach is proposed. A series of perception models are developed to decouple complex sensing tasks and assign them to specialized models. The combination of these models enables the effective solution of real-world problems.

To accurately extract specific image regions and enhance the quality of subsequent tasks, the PP-LiteSeg model\cite{peng2022pp} is employed. This lightweight semantic segmentation model incorporates a flexible decoder, a unified attention fusion module, and a simple pyramid pooling module, striking an excellent balance between accuracy and speed, making it ideal for real-time segmentation tasks. As shown in Fig.\ref{fig:model}(A), raw images captured by the Gelsight FlexiRay are fed into the PP-LiteSeg model to segment the front beam skeleton, front beam interaction region (direct perception), mirror region (reflective perception), and local contact region. The front beam interaction region and mirror region together constitute the sensing region.

Next, a proprioception model based on a ResNet50 backbone\cite{targ2016resnet} is developed, as illustrated in Fig.\ref{fig:model}(B). This model takes as input the segmented images of the front beam skeleton and the sensing region provided by the segmentation model. By using fully connected layers, the model estimates the contact normal force, the 3D position of the contact point (relative to the reference coordinate system), and the side beam node positions through the force head, position head, and node head. This model enables proprioceptive sensing of force, contact position, and structural deformation. 

Additionally, a texture perception and recognition model is built using the ResNet50 backbone, as illustrated in Fig.\ref{fig:model}(D). This model processes the sensing region image obtained from the segmentation model and is linked to a classification head composed of fully connected layers. The model outputs category labels based on the recognized contact texture.

Finally, the local contact region images extracted by the segmentation model allow for more precise detection of the interaction area’s location and details. The color information within the contact area markers is leveraged by a data-driven temperature perception model to provide temperature feedback, as illustrated in Fig.\ref{fig:model}(C). Furthermore, the relative positions of the markers across frames facilitates the extraction of slip information during contact.

\begin{figure*}[ht]
    \centering
    \includegraphics[width=1\linewidth]{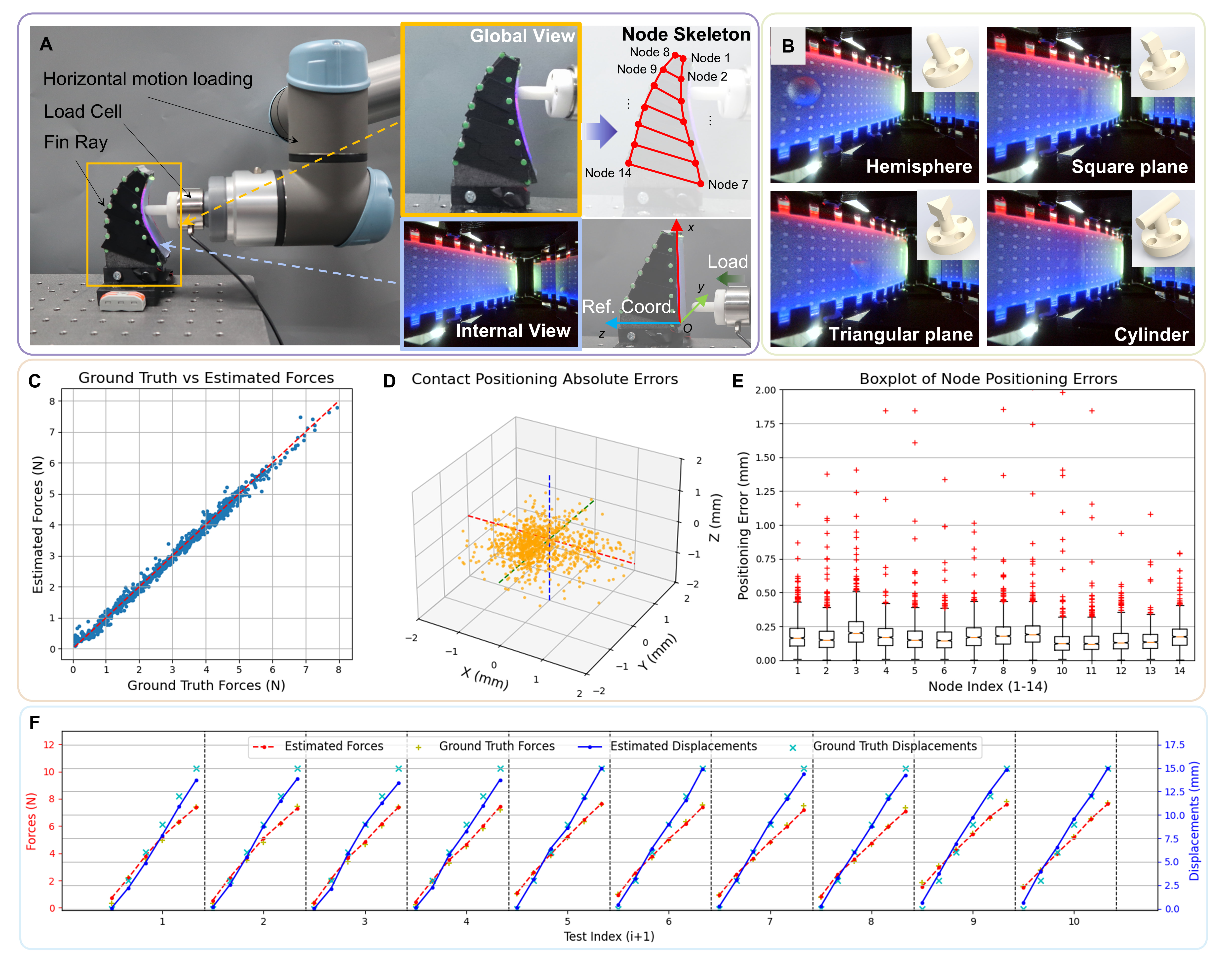}
    \caption{Force and proprioceptive perception experimental process and results. (A) Data collection platform. (B) Probe type of the load cell. (C) Accuracy analysis of force estimation. (D) Absolute error distribution of contact position prediction. (E) Box plot of positioning errors for joint nodes. (F) Continuous estimation of contact force and contact depth during dynamic contact processes.}
    \label{fig:force}
\end{figure*}

\begin{figure*}[ht]
    \centering
    \includegraphics[width=1\linewidth]{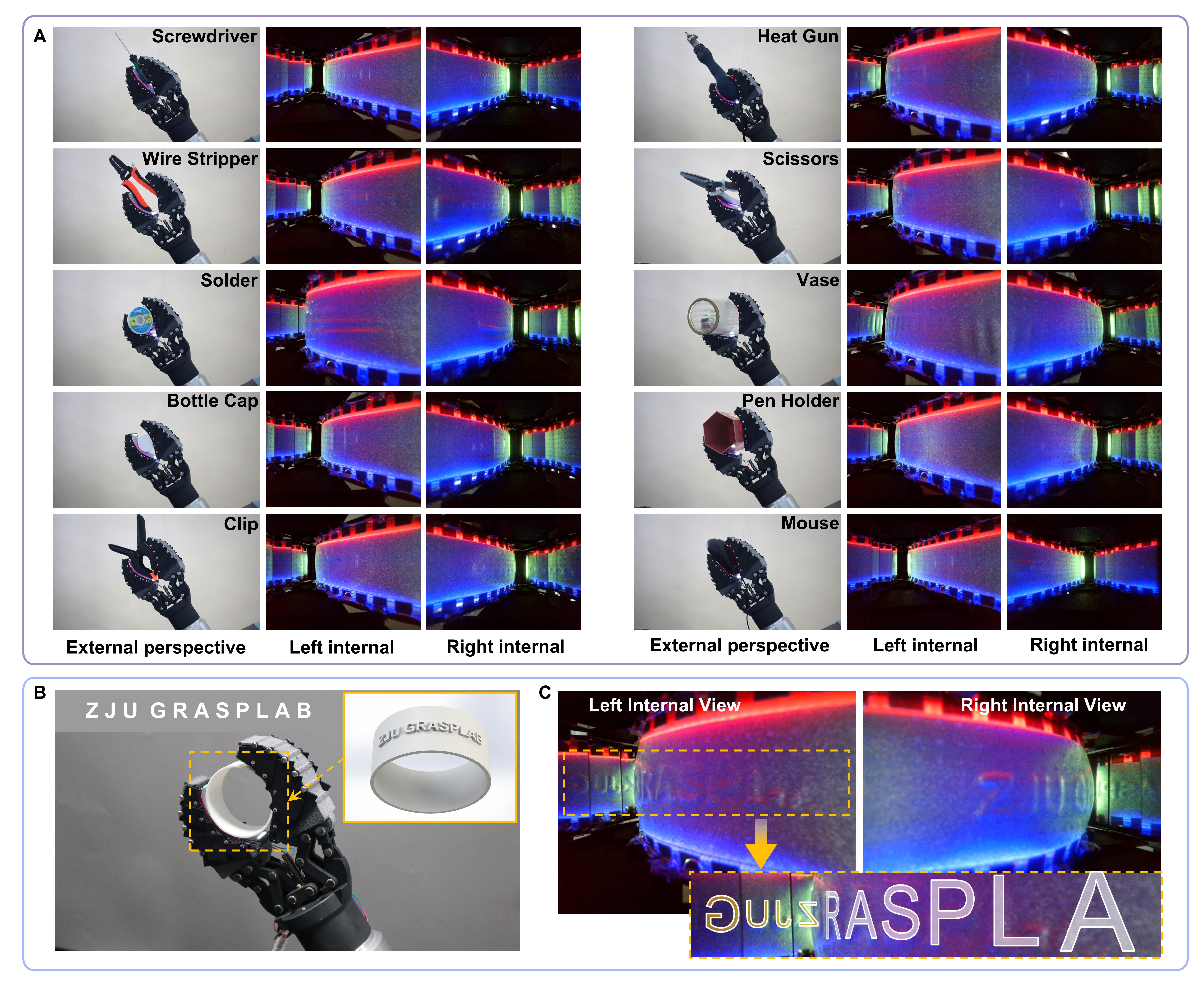}
    \caption{Texture detection performance. (A) Raw internal images captured while gripping various objects. (B) 3D-printed text ring used to evaluate fine texture detection. (C) Internal raw image of the text ring grip, illustrating texture details captured by both direct and reflective sensing.}
    \label{fig:texture}
\end{figure*}

\section{Experiments}

Drawing a parallel to the proprioceptive and discriminative touch capabilities of the human hand, this section presents experiments to evaluate the performance of GelSight FlexiRay in key sensory modalities. First, experiments on contact force, contact position, and spatial deformation perception assess FlexiRay’s proprioceptive ability to sense external forces and its own deformation. Next, tactile-based ball classification experiments test FlexiRay’s capability to perceive and recognize surface details of objects. Finally, human-robot interaction experiments with a water cup further validate FlexiRay’s temperature sensing ability, akin to human responses to thermal stimuli, as well as its ability to detect relative slippage between the finger and the object. Together, these experiments demonstrate FlexiRay's discriminative touch capabilities in recognizing objects and their properties, and assess its effectiveness and potential for practical multi-task applications.

\subsection{Proprioceptive Perception}

\subsubsection{Data collection and model training}
The dataset collection platform for proprioception tasks is shown in Fig.\ref{fig:force}(A). The Gelsight FlexiRay is mounted at a fixed position on the optical platform, with the reference coordinate system established at the lower-left corner of the tactile sensing pad in its initial state. A normal force sensor is attached to the end of the UR5e robotic arm, with the arm's tool coordinate system adjusted so that its z-axis is parallel to the z-axis of the reference coordinate system, allowing a force to be applied along this direction. A 3D-printed contact probe is attached to the force sensor, designed to simulate four contact surface types: hemispherical, square, triangular, and cylindrical, as shown in Fig.\ref{fig:force}(B). These designs replicate various contact forms, including point, edge, and flat surface interactions. Initially, the probe does not contact the finger. Random initial x and y positions within the feasible contact domain are selected, and a force is applied along the z-axis. Upon detecting contact via the force sensor, synchronized data collection is initiated. This process involves capturing side beam node images using a global camera, recording internal finger images through the embedded camera, and logging force sensor readings. Simultaneously, the robot’s end-effector positions are recorded in real-time to ensure comprehensive data alignment.

In terms of model implementation, each head of the proprioception model employs fully connected layers with 512 and 256 neurons, and the output layers have dimensions of \( \mathbb{R}^1 \), \( \mathbb{R}^3 \), and \( \mathbb{R}^{28} \), respectively. The loss functions for all three outputs are set to MSE loss, with a learning rate of 0.001. The dataset consists of 5,000 images, which are split into training and validation sets with a 4:1 ratio. Additionally, the PP-LiteSeg semantic segmentation model is built upon the STDC2 backbone. Cross-Entropy Loss is used as the loss function. The model is optimized using SGD with a momentum of 0.9 and a weight decay of $4.0 \times 10^{-5}$. A Polynomial Decay learning rate scheduler is employed, with an initial learning rate of 0.01 and a decay factor of 0.9. The performance of the trained model is evaluated, achieving a mean Intersection over Union (mIoU) of 89.98\%.

\subsubsection{Results}

Analysis of the model's accuracy in estimating force, contact point position, and measurement node position is presented below. Figure \ref{fig:force}(C) shows a scatter plot comparing the ground truth forces and the estimated forces, where the x-axis depicts the ground truth forces, and the y-axis represents the estimated forces. The red dashed line illustrates the ideal perfect match line (\(y = x\)), indicating perfect agreement between the estimated and true values. The data presented in the figure shows that the root mean square error (RMSE) of the estimated forces is 0.135 N, with a correlation coefficient of 0.997. This indicates a very strong positive correlation between the ground truth forces and the estimated forces. These results suggest that the model's estimated forces are highly consistent with the actual measured values, demonstrating a high level of accuracy in the force estimation task.

Figure \ref{fig:force}(D) illustrates the distribution of absolute errors for the estimated contact positions in the 3D space, with scatter points closer to the origin indicating smaller errors. The plot clearly shows that the majority of the errors are concentrated within a 1 mm range, suggesting that the model’s prediction errors are generally small. Statistical results reveal a mean error of 0.83 mm and a standard deviation of 0.37 mm, further validating the model's accuracy in estimating contact positions. Additionally, the density of the error distribution reflects the model's robustness in predicting contact positions across different scenarios.

For the Fin Ray, the positions of the side beam joints serve as critical indicators of the finger's deformation. These positions enable the robot to interpret the physical configuration of the finger, facilitating both interaction feedback and motion control. Figure \ref{fig:force}(E) presents a boxplot of positioning errors for the 14 nodes of Gelsight FlexiRay, illustrating the estimation accuracy for each node in the contact localization task. The results indicate that the average positioning errors for most nodes are relatively small, around 0.19 mm, with an average standard deviation of approximately 0.37 mm. Nodes 11 and 12 show notably larger positioning errors, with mean errors of 0.1710 mm and 0.1863 mm, respectively, and standard deviations significantly higher than the other nodes (0.6897 mm and 0.8175 mm). These two nodes are located on the lower half of Gelsight FlexiRay's back beam, where greater deformation occurs during interaction. Additionally, the images captured by the camera primarily focus on the surface of the rigidly connected mirror, without capturing the deformation of the actual back beam, which likely contributes to the higher positioning errors observed at these nodes. Overall, however, the model demonstrates high positioning accuracy and robustness, effectively meeting the localization requirements for proprioceptive perception.

To evaluate perceptual accuracy during continuous interaction, the probe transitions from a randomly initialized contact position to the target deformation position, with data on the normal force and contact depth along the z-axis of the reference coordinate system being collected simultaneously. The comparison between the real measurement data and predicted results from ten repeated interactions is shown in Fig.\ref{fig:force}(F). It is evident that the model maintains high accuracy and stability in estimating both force and contact depth under dynamic, continuous prediction. Notably, at a contact force of 7.5 N, the deformation in the contact depth direction is approximately 15 mm. In comparison, existing sensors of this type, such as the GelSight Baby Fin Ray\cite{liu2023gelsight}, show a deformation of around 3-4 mm at 7.5 N contact force, indicating that the compliance of the GelSight FlexiRay is more than four times higher than that of the most advanced sensors in this category.

\subsection{General Texture Detection Performance}

To assess the texture detection performance of Gelsight FlexiRay, a range of large curved or wide-area gripped objects are selected. The objects include tools like screwdrivers, hot air guns, wire strippers, and scissors. Furthermore, curved items of different sizes and shapes, such as solders, bottle caps, vases, pencil holders, clips, and mice, are also tested. A two-finger gripper composed of Gelsight FlexiRay is mounted on a UR5e robotic arm to perform natural grasping experiments and capture texture images without external interference. The raw images captured by the internal cameras of the left and right fingers are shown in Fig.\ref{fig:texture}(A). The results demonstrate that Gelsight FlexiRay not only conforms seamlessly to and wraps around large curved objects but also accurately captures the surface contours and geometric details of each object, even during flexible deformations. For example, in the case of the vase, Gelsight FlexiRay records uniformly distributed surface stripes on the sensing area, with clear contact textures captured in the fingertip region of the mirrored area. Overall, Gelsight FlexiRay provides complete coverage of the contact area, showcasing a unique large-area tactile perception capability that is not available in current VTS.

Additionally, Gelsight FlexiRay's performance and fine texture detection capability are evaluated using a 3D-printed text ring shown in Fig.\ref{fig:texture}(B). The ring has an outer diameter of 68 mm, a font height of 1.5 mm, a line width of 1 mm, and no chamfering. As shown in the internal view in Fig.\ref{fig:texture}(C), under large deformations, the front beam of Gelsight FlexiRay obstructs the fingertip region, preventing the camera from capturing a complete image of the contact area. However, by incorporating mirror-reflective regions that passively adjusts with deformation, Gelsight FlexiRay can capture the occluded texture, thus providing a more complete perceptual field. In the internal view of the left finger, the direct perception area captures "RASPLA," while the reflective perception area clearly captures "ZJUG," forming a continuous texture pattern "ZJUGRASPLA" within the contact range. This further confirms the effectiveness of the designed direct and reflective sensing strategies, along with the optimization methods.

\subsection{Texture-Based Ball Classification}

\begin{figure*}[ht]
    \centering
    \includegraphics[width=1\linewidth]{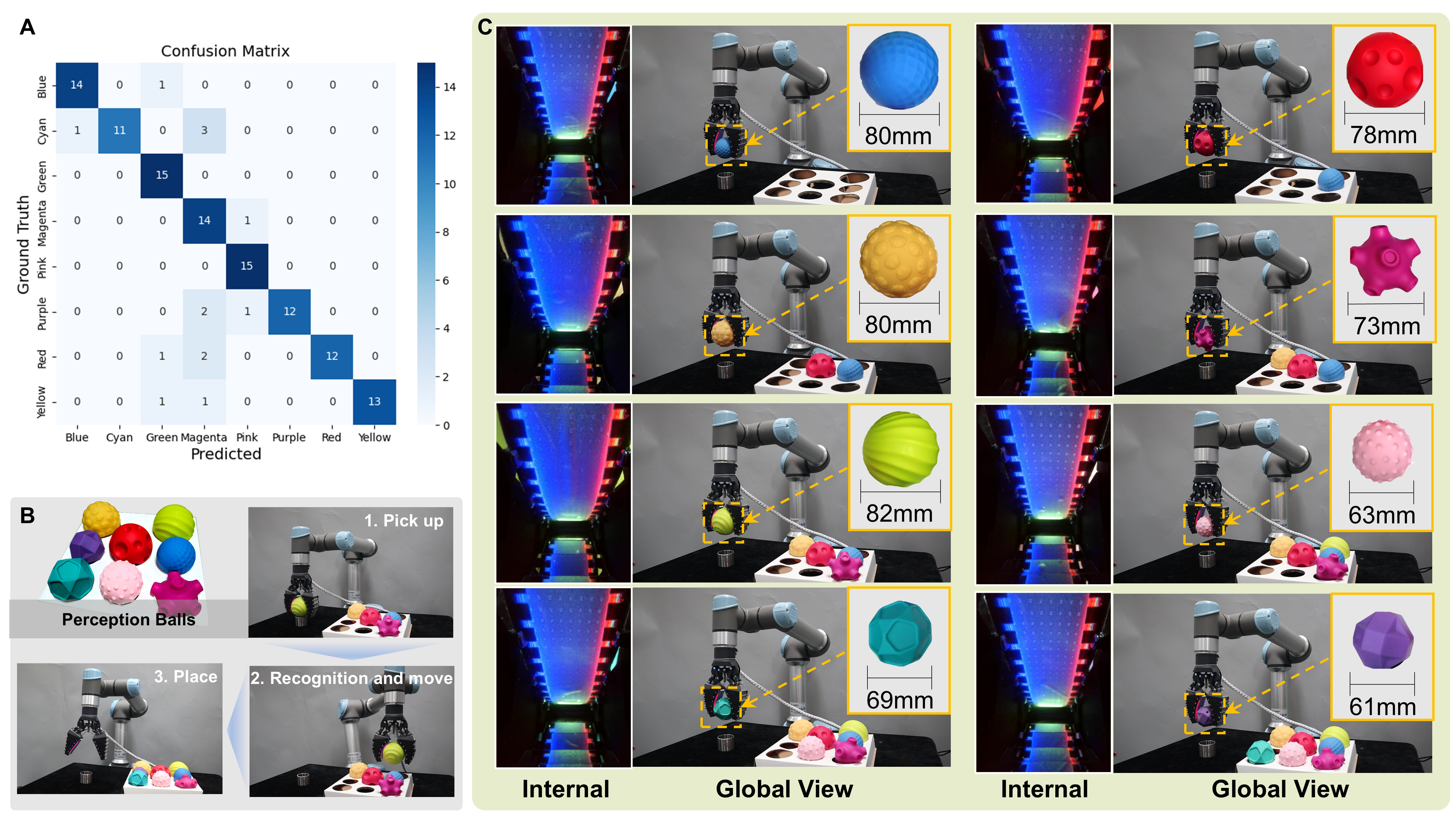}
    \caption{Tactile-based ball classification experiment process and results. (A) Confusion matrix of the test results. (B) Workflow for ball sorting and tactile perception classification. (C) Raw tactile images of various ball types.}
    \label{fig:ball}
\end{figure*}

\subsubsection{Data collection and model training}

To demonstrate the recognition and classification capability of Gelsight FlexiRay on texture, eight perception balls with different surface textures are selected as classification targets. The grasp position and force are manually adjusted, and a total of 600 tactile images of stable grasps in various random states are collected. The dataset is split into a training set and a validation set in a 4:1 ratio, with data augmentation applied. The classification head of the recognition model, consisting of fully connected layers with 512 and 256 neurons, outputs the predicted category label. The loss function used is Cross-Entropy Loss, and the training process employs the Adam optimizer with a learning rate set to 0.001.

\subsubsection{Results}

The classification model achieves an accuracy of 95.83\% on the manually collected validation dataset. To assess performance under realistic grasping conditions, each ball is randomly grasped 15 times, resulting in a test set of 120 images. The confusion matrix of the test results is shown in Fig.\ref{fig:ball}(A). The average success rate is 88.33\%, with the highest accuracies of 100\% for the green and pink balls, and the lowest accuracy of 73.3\% for the cyan ball. Analysis of the misclassifications reveals that most errors involving the cyan ball are due to interference from the magenta ball. This is likely caused by the similarity in the contact edges of the two balls, especially under light grasping conditions, which led to similar edge features in the tactile images.

Leveraging the trained classification model, a robotic ball sorting task is performed. During the experiment, various types of balls are randomly placed at the grasping position. The robot relies solely on tactile modality to perceive the surface textures of the balls, classifying them and placing them into predefined locations based on the texture features. Figure \ref{fig:ball}(C) presents tactile image frames from the grasping process, where the surface texture features of the different balls are clearly distinguishable, highlighting the sensor's ability to capture texture information during real-world grasping tasks.

The experimental results demonstrate that the classification model effectively uses tactile image data to accurately identify and classify the surface textures of different balls. Moreover, the gripper's performance in the sorting task reflects the overall compliance and stability of the Gelsight FlexiRay gripper, allowing them to adapt to various surface shapes and textures. The robot's ability to precisely classify the balls based solely on tactile perception underscores the importance of tactile modality in precise grasping and manipulation. This experiment further validates the effectiveness and practicality of Gelsight FlexiRay and the classification model in real-world applications.

\subsection{Temperature Sensing and Human-Robot Cup Transfering}

\begin{figure*}[ht]
    \centering
    \includegraphics[width=1\linewidth]{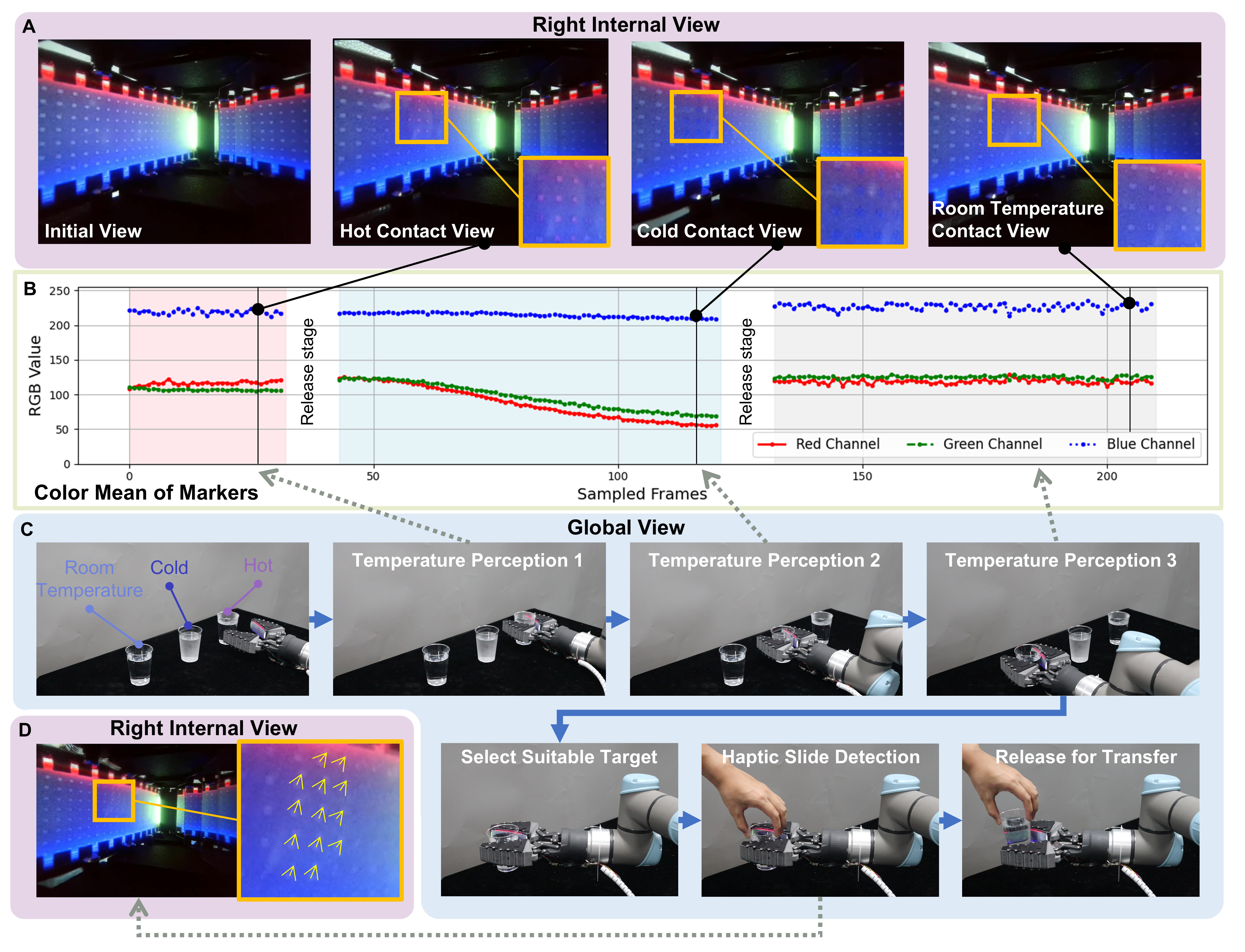}
    \caption{Tactile temperature sensing and sliding detection for human-robot cup interaction. (A) Raw tactile images. (B) Changes in the RGB values of tactile perception area markers under different temperature contacts. (C) Robotic cup grasping and temperature sensing process. (D) Sliding signal detection.}
    \label{fig:water}
\end{figure*}

\subsubsection{Experimental procedure}

Humans perceive temperature through touch, utilizing thermoreceptors in the skin to detect thermal stimuli and differentiate between various temperatures. Additionally, tactile feedback enables the hands to naturally sense external forces acting on held objects, allowing for precise handling and transfer. Drawing inspiration from these human capabilities, this experiment evaluates the Gelsight FlexiRay gripper’s ability to replicate similar sensory functions. Using a water cup transfer scenario, the robot assists in selecting a cup at the desired temperature and facilitates its handover to a human. The experiment focuses on assessing Gelsight FlexiRay’s temperature perception and tactile feedback mechanisms, investigating how temperature sensing identifies target objects and how tactile feedback controls release timing to enable efficient and safe interactions. The experimental procedure is outlined as follows:

\begin{itemize}
    \item \textbf{Tactile image acquisition of water cups:} 
    At the start of the experiment, three cups of water are placed on a table: hot water (80°C), cold water (4°C), and room temperature water (24°C). The cups are unmarked, and their temperatures cannot be visually distinguished without physical contact. The robot grasps each of the three cups while capturing sensing images from Gelsight FlexiRay.
    
    \item \textbf{Temperature perception and recognition:}
    The contact area of each cup is segmented, and the marker pixels within this region are extracted to compute the average RGB values. The temperature recognition model employs a dual fully connected layer architecture with an input dimension of 256x256, producing output labels to classify the water as hot, cold, or at an optimal temperature. The perception process is maintained for a sufficient duration to allow heat transfer to stabilize, ensuring a reliable temperature modality image.
    
    \item \textbf{Cup transfer and release triggering:} 
    During the cup transfer process, when a human hand contacts the cup and attempts to take it from the robot’s grasp, a sliding signal is generated in the tactile image. Gelsight FlexiRay detects the slip amount in real-time from the contact area. Once predefined conditions are met, the release mechanism is activated, automatically loosening the gripper to allow the human to safely take the cup.
\end{itemize}

\subsubsection{Results}

The tactile image frames and corresponding RGB color changes throughout the experiment are shown in Fig.\ref{fig:water}(A) and (B). The results indicate that the robot accurately recognizes the temperature of the water cups in a dynamic environment and successfully delivers the appropriate cup based on the human’s preference. During the interaction, Gelsight FlexiRay reliably detects the sliding signals triggered by human contact and releases the cup at the optimal moment. In repeated trials, the robot successfully completed the cup transfering and released actions without any incidents of dropping the cup or misjudging the temperature, demonstrating the system’s robustness and reliability.

This experiment underscores the integrated application of the flexible, large-area VTS developed in this study, highlighting its potential for the multimodal perception. It illustrates the multidimensional intelligence of robots in perception, action, and human-robot interaction, with promising applications in more complex tasks such as elderly caregiving and handling objects in hazardous environments.

\section{Conclusion}

This work presents a novel VTS-integrated soft robotic hand inspired by Finray effect, seamlessly combining enhanced structural compliance with advanced multimodal sensing capabilities. Optical path disturbances caused by large structural deformations are treated as critical design parameters. Based on this, segmented mirror arrays are introduced, and the sensor’s layout is systematically optimized. These innovations enable the FlexiRay to deliver precise, stable, and consistent multimodal sensing across force, texture, slippage, temperature, and proprioception, while maintaining the intrinsic flexibility and adaptability of the FRE structure. To achieve this, we begin by characterizing the force-deformation patterns of the FRE finger under various loading conditions in physical settings. These patterns serve as seeds for determining the optimal positioning and orientation of the mirrors. This approach ensures consistent image acquisition via controlled mirror reflections, even under substantial deformations. As a result, the FlexiRay excels in grasping objects with irregular, cylindrical, or spherical geometries, significantly increasing conformal contact areas and enriching perceptual data during compliant interactions with improved efficiency. Extensive experimental validation highlights the sensor's exceptional performance, achieving a force resolution of 0.14 N and a proprioceptive positioning accuracy of 0.19 mm. Compared to state-of-the-art compliant VTS systems, the FlexiRay demonstrates fivefold greater deformation capacity under identical loads while maintaining consistent coverage of the front contact surface. Neural network analyses further validate its capability to recognize multimodal contact information, underscoring its utility in diverse tasks such as grasping, manipulation, and object classification. These results address longstanding limitations of VTS in handling large deformations, offering a compact, high-resolution, and systematically efficient solution for tactile sensing.

The FlexiRay represents a significant step forward in advancing high-resolution, larger-coverage yet low-cost sensory perception for soft robotic systems. Its design leverages physical deformation as a functional parameter, transforming challenges into opportunities for enhanced sensing performance. Future work will focus on integrating adjustable skeletal stiffness and advanced perceptual domain materials to further refine sensory fidelity, particularly in texture perception. Additionally, we aim to extend the FlexiRay's capabilities to multi-fingered grippers and explore applications in dynamic, high-safety tasks such as fruit harvesting and assistive care, paving the way for next-generation robotic systems capable of robust, adaptable, and intelligent interactions.

% \section{Acknowledgements}

% The authors gratefully acknowledge the financial support from the National Key Research and Development Program of China (Grant no. 2023YFB4704700).

% \section{Author Disclosure Statement}

% No competing financial interests exist.

% \section{Supplementary Material}

% \par Supplementary Movie S1
% \par Supplementary Movie S2
% \par Supplementary Movie S3

% \section{Appendix}

% Here we provide detailed insights into the capabilities of tactile sensing oad in perceiving the texture of planar objects. The results are illustrated in Figure 1.

\bibliographystyle{IEEEtran}
\bibliography{reference2.bib}

\end{document}